\documentclass[a4paper,fleqn]{cas-sc}

\usepackage[authoryear,longnamesfirst]{natbib}
\usepackage{algorithm}
\usepackage{algorithmic}

\usepackage{subcaption}
\def\tsc#1{\csdef{#1}{\textsc{\lowercase{#1}}\xspace}}
\tsc{WGM}
\tsc{QE}
\tsc{EP}
\tsc{PMS}
\tsc{BEC}
\tsc{DE}

\begin{document}
\let\WriteBookmarks\relax
\def\floatpagepagefraction{1}
\def\textpagefraction{.001}

\shorttitle{Selective Cooperative Perception via Blind Spot Reasoning}

\shortauthors{Ma et~al.}

\title[mode=title]{Selective Cooperative Perception via Blind Spot Reasoning and Agentic Dialogue with Vision-Language Models}



\author[1,2]{Chengyuan Ma}[
]
\ead{cyma@nju.edu.cn}

\author[2]{Shixiao Liang}
\ead{sliang85@wisc.edu}

\author[2]{Hangyu Li}
\ead{hli939@wisc.edu}

\author[2]{Zheng Li}
\ead{zli2674@wisc.edu}

\author[2]{Hang Zhou}
\ead{hzhou364@wisc.edu}

\author[2]{Keke Long}
\ead{klong23@wisc.edu}

\author[2]{Bofeng Cao}
\ead{bcao44@wisc.edu}

\author[2]{Zhaohui Liang}
\ead{zhaohui.liang@wisc.edu}

\author[2]{Xiaopeng Li}[
]
\ead{xli2485@wisc.edu}


\affiliation[1]{%
    organization={School of Robotics and Automation, Nanjing University},
    city={Nanjing},
    state={Jiangsu},
    country={China}
}

\affiliation[2]{%
    organization={Department of Civil \& Environmental Engineering, University of Wisconsin--Madison},
    city={Madison},
    state={WI},
    country={United States}
}




\begin{abstract}
Real-time cooperative perception (CP) in multi-agent traffic environments is highly resource-intensive, which posing scalability challenges for large-scale deployment. Efficiency can be significantly improved by selectively performing CP only in safety-critical scenarios and for key occluding objects. Vision-Language Models (VLMs) are well-suited for complex reasoning in dynamic and uncertain environments, making them a promising tool for this selective CP task. This study proposes a VLM-driven agentic framework for blind spot reasoning and dialogue-based CP. A mixed-model architecture is designed that integrates conventional object detection for stable perception, vision-language reasoning for identifying impactful occlusions, and a physics-based geometric module to align reasoning with real-world visibility constraints. The framework also incorporates a Chain-of-Thought (CoT) prompting mechanism to enhance interpretability and intent-aware decision-making. We collect real-world data from connected vehicles operating in urban traffic and compare the driving-critical blind spot identification and dialogue performance of four state-of-the-art VLMs. Experimental results demonstrate that our method achieves over 80\% accuracy in driving-critical blind spot detection and over 90\% success in dialogue-driven perception sharing. This work provides a scalable, interpretable, and bandwidth-efficient alternative to conventional CP pipelines, and offers new insights into agentic communication design for future connected traffic systems.
\end{abstract}



\maketitle
\section{Introduction}

Cooperative Perception (CP) plays a crucial role in enhancing vehicle situational awareness by enabling multiple agents to share perceptual information~\citep{bai2024survey}. This capability allows vehicles to overcome limitations caused by occlusions, narrow sensor fields-of-view, or hardware failures. In complex traffic environments, especially those involving high-speed interactions or obstructed views, CP has the potential to significantly improve driving safety and decision-making accuracy by complementing ego perception with external observations \citep{yuan2023enhanced}.

However, in traffic scenarios with numerous connected agents, the cost of exchanging perceptual information can quickly become prohibitively high. Traditional fusion-based CP methods are commonly categorized into early, intermediate, and late fusion~\citep{bai2024survey}. Early fusion aggregates raw sensor data such as point clouds or images, providing rich scene representations but requiring extremely high bandwidth and strict temporal synchronization~\citep{bai2022pillargrid}. Intermediate fusion exchanges processed features or object-level proposals, reducing bandwidth but remaining highly sensitive to calibration errors~\citep{xu2022v2x}. Late fusion transmits only final perception outputs, minimizing data size but offering limited robustness to occlusions, as missing information cannot be recovered. Although effective in controlled settings, these approaches face increasing limitations in real-world traffic, where many agents must continuously exchange high-frequency perception updates with multiple peers. The resulting bandwidth and computational burden hinders the scalability of conventional fusion-based CP. Recent works have highlighted these challenges and emphasized the need for more bandwidth-efficient CP alternatives~\citep{bai2024survey, bai2023cooperverse}.

This motivates a shift toward selective CP. In most driving scenarios, vehicles do not require external information and can operate safely based on ego perception alone~\citep{ma2026realtime}. CP becomes essential only in certain contexts such as during occlusions, complex interactions, or ambiguous intent prediction~\citep{zheng2024intelligent, liu2025sracp}. Therefore, if a system can reason about when CP is necessary, from whom, and what specific information is needed, it becomes possible to perform targeted CP with minimal cost. Prior studies have proposed early forms of this concept, including rule-based filters and context-aware triggers~\citep{bai2023cooperverse}. However, due to the dynamic and highly contextual nature of traffic, such rule-based approaches struggle to generalize across diverse real-world scenarios~\citep{zhang2023building}.

To enable reasonable judgment of when CP is necessary across diverse environments, the perceptual and reasoning capabilities of VLMs provide a promising solution. These VLMs can reason over complex logical relationships by integrating scene understanding, driving intentions, object-level dynamics, and commonsense priors~\citep{xu2024vlm, jiang2025survey}. Such reasoning capabilities are well-suited to determining when CP is required and formulating targeted queries for specific perceptual information. Recent work has explored using VLMs in agent communication, intention prediction, and modular perception–planning loops~\citep{zhang2023building, zhang2024proagent, chiu2025v2v, du2024constrained}. These studies demonstrate the potential of VLMs to perform decision-making with reduced communication overhead, but further investigation is needed to align these capabilities with safety-critical constraints.


Despite their promise, current VLM-based approaches face two major limitations that hinder their direct application to safety-critical perception. First, their visual understanding often suffers from perceptual hallucination, where the model misidentifies or fabricates objects due to weak grounding between visual tokens and real-world geometry~\citep{huang2025survey}. This undermines reliability in tasks such as occlusion reasoning and obstacle detection, VLMs exhibit limited capability in spatial and numerical reasoning. They struggle to maintain geometric consistency and accurately infer object positions or trajectories in metric space, as their reasoning is primarily symbolic rather than physics-grounded. Consequently, purely VLM-driven pipelines \citep{du2024constrained} remain prone to high spatial errors (often exceeding 50\%) and hallucinations \citep{chiu2025v2v} in coordinates understanding or trajectories prediction tasks, posing unacceptable risks in real-world driving. 

To address these limitations, this study proposes a hybrid framework that combines the robustness of conventional computer vision models, the reasoning and interaction capability of VLMs, and the consistency of physics-based geometric models. In this framework, a computer vision-based detector extracts object types and positions with stable numerical accuracy, avoiding hallucinations. VLMs are then used to identify which occlusions constitute driving-critical  blind spots based on driving intention. Finally, an agentic dialogue mechanism is used to selectively query peer agents for supplementary perception in these blind spots. The contributions of this work are twofold: 
(1) a VLM-based blind-spot reasoning module that incorporates driving intent to identify high-impact occlusions, together with a physics-grounded geometric visibility verifier that aligns VLM outputs with real-world occlusion constraints; and
(2) an agentic dialogue framework that enables selective CP with minimal communication cost while maintaining safety in critical scenarios.

The remainder of this paper is organized as follows. Section~\ref{sec:method} presents the proposed methodology, including the mixed-model architecture for blind-spot reasoning and dialogue-based selective CP, as well as its integration with geometric modeling. Section~\ref{sec:experiment} details the experimental validation using real-world data collected from connected vehicles operating in urban traffic scenarios. Section~\ref{sec:conclusion} concludes the study and outlines future research directions.


\section{Methodology}
\label{sec:method}

\subsection{Framework Overview}

\begin{figure}
    \centering
    \includegraphics[width=1\linewidth]{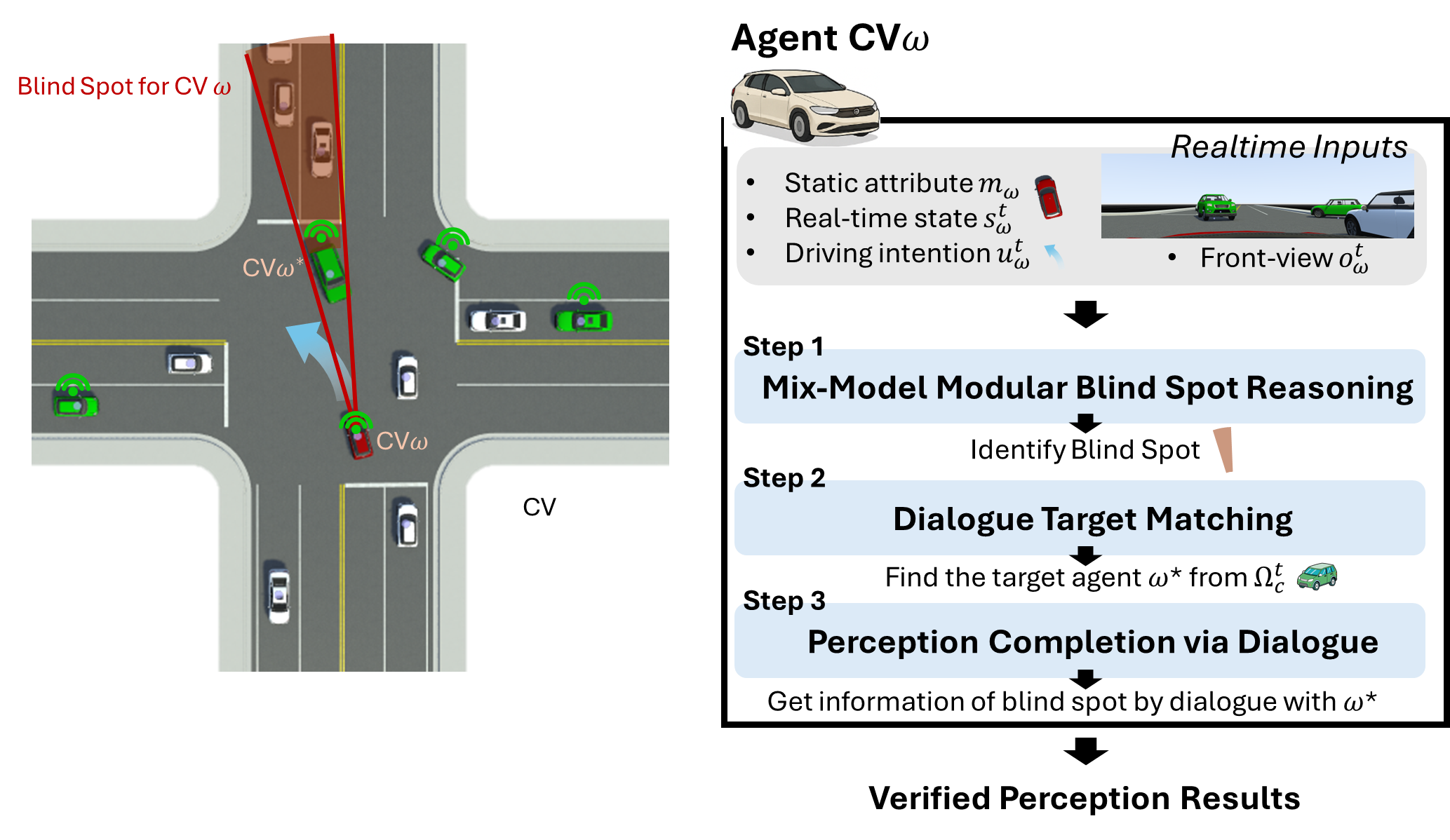}
    \caption{Overview of the Agentic AI-based Selective Cooperative Perception Framework}
    \label{fig:framework}
\end{figure}

Figure~\ref{fig:framework} illustrates the proposed blind spot reasoning and selective CP scenario. We consider a discrete-time mixed traffic environment over time steps \( t \in \mathcal{T} := \{0, 1, \ldots, T\} \), composed of both connected and non-connected agents (shown as colored and white participants in Figure~\ref{fig:framework} , respectively). Let \( \Omega^t \) denote the set of all agents at time \( t \), partitioned into connected vehicles (CVs) \( \Omega^t_C \) and non-connected vehicles (NVs) \( \Omega^t_N \), i.e., \( \Omega^t = \Omega^t_C \cup \Omega^t_N \). Each connected agent \( \omega \in \Omega^t_C \) is equipped with V2X communication capabilities, enabling information sharing and dialogue-based coordination with other CVs. For each CV \( \omega \), we define four key components of its internal state:  
(1) \textbf{Static attribute} \( m_\omega \), which includes appearance features (e.g., color), semantic class \( z_\omega \) (e.g., light-duty vehicle, pedestrian), and physical dimensions \( (l_\omega, w_\omega) \);  
(2) \textbf{Real-time state} \( s_\omega^t \), including the center position \( (x_\omega^t, y_\omega^t) \), velocity \( v_\omega^t \), and heading angle \( \phi_\omega^t \);  
(3) \textbf{Driving intention} \( u_\omega^t \), expressed in natural language (e.g., "turning left at the intersection" as in the case of CV \( \omega \) in the figure);  
(4) \textbf{Perception observation} \( h_\omega^t \), referring to onboard sensor input. In this study, we focus specifically on the \textbf{front-view} image \( o_\omega^t \), which serves as the basis for blind spot analysis. This formulation naturally generalizes to other camera views.
Under normal conditions, at each time step, each CV agent makes driving decisions based solely on its own onboard perception observation \( h_\omega^t \). Building on this baseline, this study introduces a supplementary module to assess whether the agent’s individual forward-facing perception is sufficient for safe and informed decision-making. If not, the system actively supplements the perception through communication with other connected agents. 

Therefore, the goal of this study is for each CV to autonomously reason about driving-critical blind spots that may affect its driving safety and, when necessary, selectively acquire information from other CVs with minimal communication overhead. To balance perception completeness and communication efficiency, we propose an agentic framework for selective CP. From an information-sharing perspective, the static attribute \( m_\omega \) and real-time state \( s_\omega^t \) are lightweight and periodically broadcast to all CVs to maintain a common situational understanding. In contrast, the perception observation \( h_\omega^t \) is high-volume and therefore unsuitable for continuous transmission under limited bandwidth.

Each CV \( \omega \) independently determines whether CP is necessary. Step 1 involves driving-critical blind spot inference using its front-view image \( o_\omega^t \) and driving intention \( u_\omega^t \). We adopt a \textbf{Mix-Model Modular Blind Spot Reasoning} architecture that integrates conventional computer vision accuracy with the reasoning capacity of VLMs. The reasoning process is structured as a CoT composed of scene understanding, object detection, intent interpretation, and blind spot analysis, allowing modular and robust inference of potential occlusions.

If no blind spot is identified or if the detected blind spot is deemed non-critical for the current driving task, the agent skips Step 2 and Step 3 and directly proceeds with its own perception output as sufficient. In most scenarios, this allows the system to operate without requiring additional communication or dialogue, thereby preserving bandwidth.

If a driving-critical blind spot is identified, the system proceeds to Step 2: \textbf{Dialogue Target Matching}. Using the shared \( m_\omega \) and \( s_\omega^t \) of all CVs in the detection area, the ego agent identifies which peer is most likely to observe the occluded region. To avoid hallucinations commonly seen in direct VLM-based spatial reasoning, the ego agent uses a physics-based geometric model that determine the most appropriate dialogue target based on relative positions and orientations..

Finally, Step 3: \textbf{Perception Completion via Dialogue} is initiated. The ego CV engages in targeted, context-aware dialogue with the selected peer rather than broadcasting its entire perception. This agentic communication framework enables CVs to cooperatively resolve perception gaps in a highly efficient manner, preserving bandwidth while maintaining safety and fluid traffic operations.

This targeted and conditional communication approach not only eliminates potential risks but also reduces the overhead compared to general CP frameworks by engaging in information exchange only when necessary.

\subsection{Mix-Model Agentic Blind Spot Reasoning}
 
\begin{figure}
    \centering
    \includegraphics[width=1\linewidth]{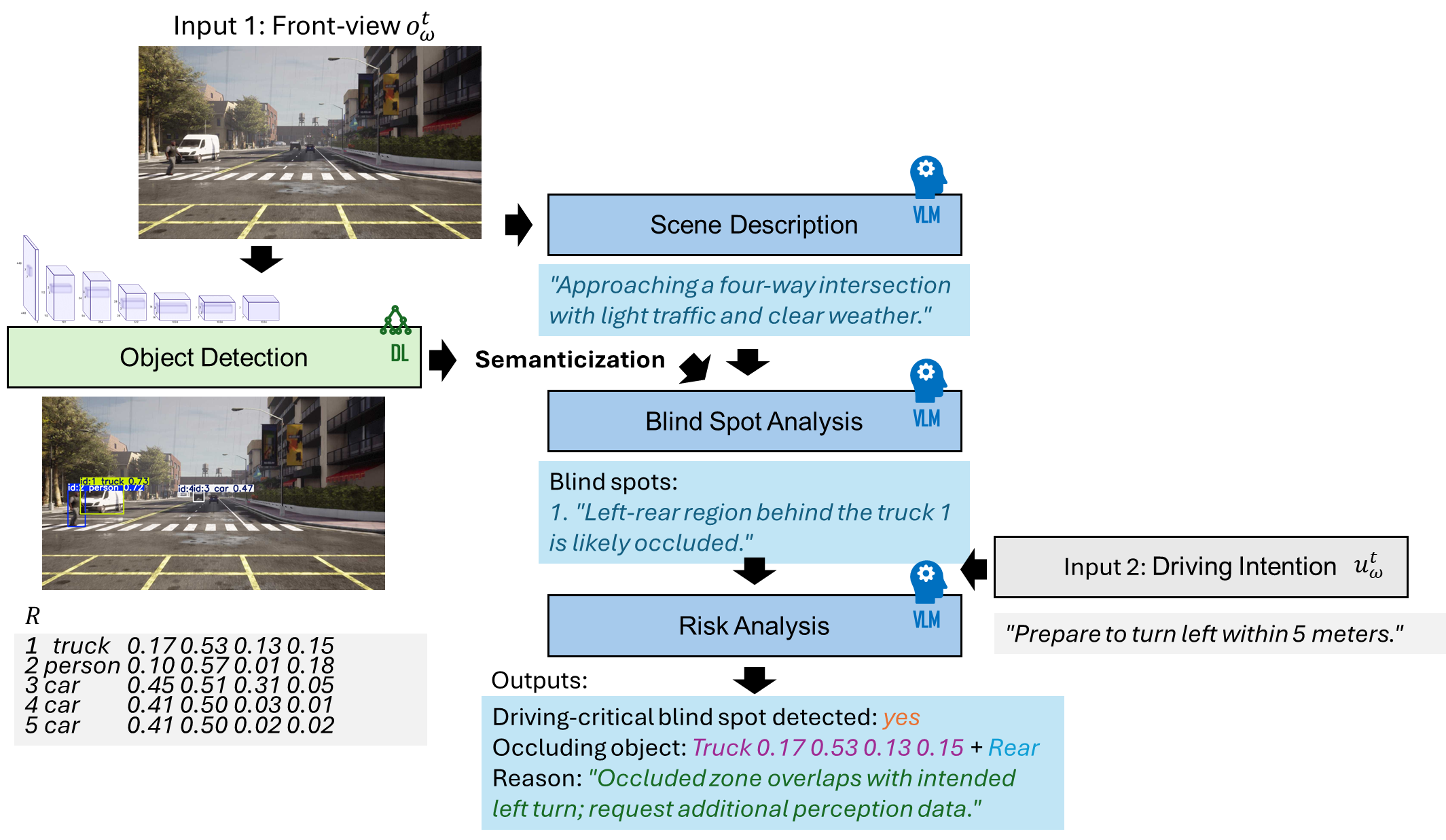}
       \caption{Modular Chain-of-Thought reasoning structure. }
    \label{fig:perception_chain}
\end{figure}

This section introduces how a CV agent determines whether CP is needed. We propose a mixed-model structure that incorporates a designed CoT agentic mechanism, as illustrated in Fig.~\ref{fig:perception_chain}. At each time step \( t \), the input to agent \( \omega \in \Omega_C^t \) includes its front-view image \( o_\omega^t \) captured by its onboard camera, and its driving intention \( u_\omega^t \), represented as a high-level context prompt. The model outputs include three key elements: whether a \textit{driving-critical blind spot} is detected (if not, Steps 2 and 3 are skipped), the corresponding \textit{occluding object}, and the \textit{reason} for the driving-critical blind spot determination.

The CoT agentic mechanism includes three main steps: (1) scene understanding, (2) detection of blind spots within the scene (without considering driving intent), and (3) identification of driving-critical blind spots by incorporating the ego vehicle’s intention. All three steps are carried out using the VLMs. However, prior studies and our experiments consistently show that VLMs perform poorly in spatial and geometric reasoning when applied directly to raw images, due to their weak geometric grounding~\citep{tian2025nuscenes, tang2024sparkle}. To improve reliability, we first apply a conventional deep learning (DL) model to detect and classify objects in the scene, as these models offer more stable performance in object recognition. The detected objects are then provided to the VLMs to assist in higher-level reasoning and judgment. Since the DL model outputs precise numerical data (e.g., bounding boxes, categories), which are not ideal for direct VLM input, we introduce a semantic conversion process termed \textbf{semanticization}. This step transforms raw detection outputs into structured natural language descriptions that are more compatible with VLM input. The specific functions of each module are described below:

At first, the agent performs a \textit{Driving Scene Description} step, extracting high-level contextual features from the input front-view image \( o_\omega^t \), such as road layout, intersection presence, lane boundaries, traffic signals, and weather conditions. These environmental cues provide essential context for interpreting visual content and anticipating traffic dynamics.

Next, the \textit{Object Detection} module identifies all visible entities, including surrounding vehicles, pedestrians, and static infrastructure. While VLMs can be used for object-level semantic parsing, they often suffer from hallucinations or unreliable bounding box precision in complex traffic scenes. To ensure accurate localization, we first apply a DL–based object detector (e.g., YOLO \citep{Jocher2023YOLO}) to obtain structured bounding boxes for all objects in the front-view image \( o_\omega^t \). For each detected object \( i \), the model outputs its class label \( z_i \), a unique identifier, and a rectangular bounding region \( R_i = \{r_i^{\mathrm{lt}}, r_i^{\mathrm{lb}}, r_i^{\mathrm{rt}}, r_i^{\mathrm{rb}}\} \), where each corner point is defined in image-space proportion (normalized coordinates), as illustrated in Fig.~\ref{fig:perception_chain}.

These bounding boxes are difficult for the VLMs to interpret directly in numerical form. To bridge this gap, we perform a semanticization step that transforms raw visual outputs into discrete textual descriptions. While object categories \( z_i \) are already suitable for textual understanding, the object position and size require semantic transformation.

\begin{figure}
    \centering
    \includegraphics[width=0.7\linewidth]{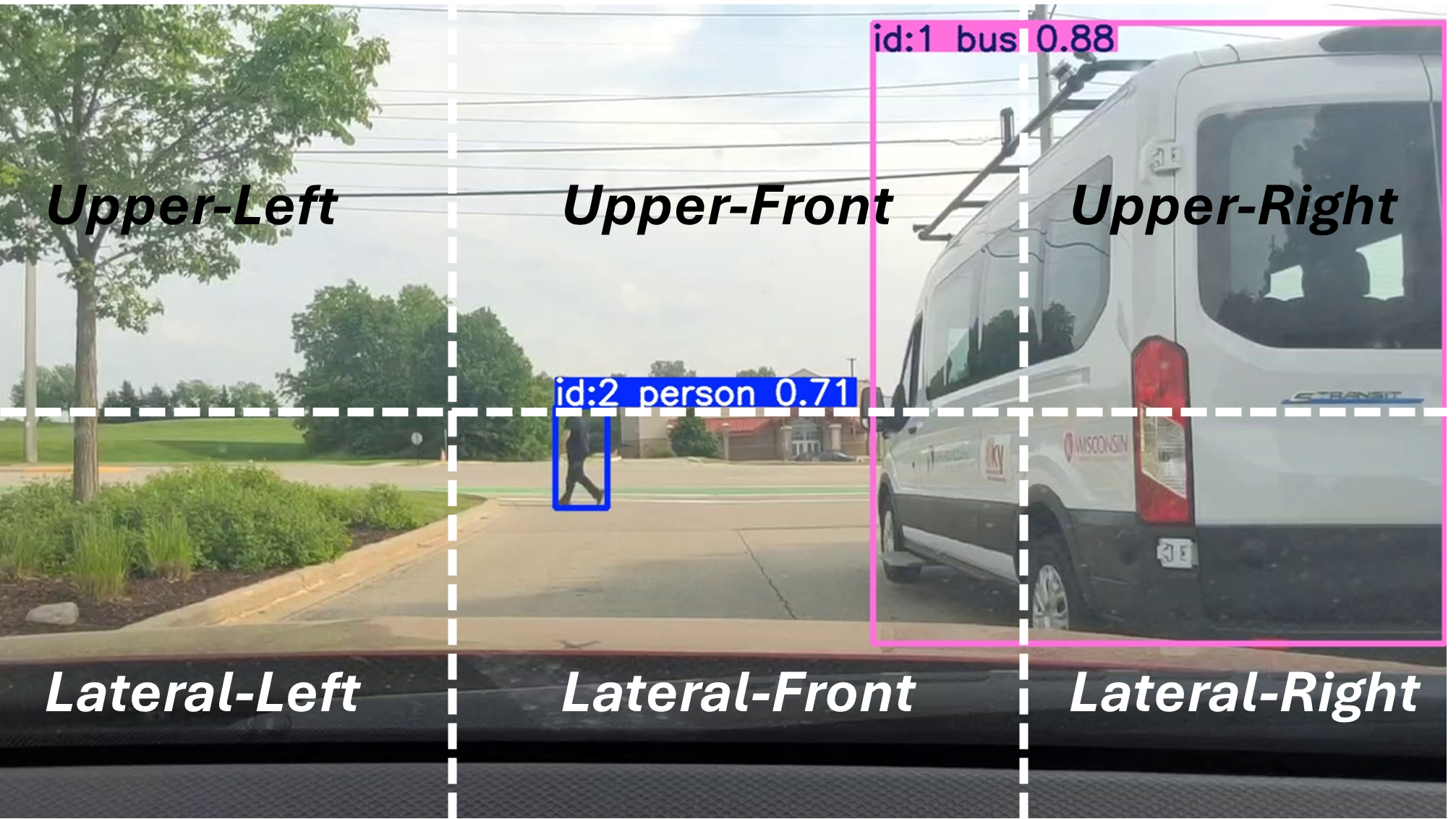}
    \caption{Semantic Region Mapping of Image Coordinates}
    \label{fig:Seman}
\end{figure}

We divide the image space into six semantic regions based on the center location of each bounding box: Lateral-Left, Lateral-Front, Lateral-Right, Upper-Left, Upper-Center, and Upper-Right, as shown in Fig.~\ref{fig:Seman} . This partitioning scheme is inspired by the explainability-oriented design in~\cite{hou2025explainable}.

Object size is also mapped to semantic categories based on the proportion of the image occupied by its bounding box. Specifically, objects whose bounding area exceeds 30\% of the total image are labeled as \textit{large}, those below 5\% as \textit{small}, and all others as \textit{medium}. These semantic descriptors—category, position, and size—serve as inputs to the VLMs to support more stable and interpretable reasoning.

Based on the results of scene understanding and object recognition, the \textit{Blind Spot Analysis} module evaluates occlusion and visibility uncertainty by reasoning over the spatial layout of detected objects. The primary goal of this module is to identify physical obstructions—such as large vehicles blocking adjacent lanes—that may conceal other agents or infrastructure behind them. At this stage, the analysis is limited to visual occlusions and does not consider the ego vehicle’s driving intention.

This reasoning process is handled by the VLMs, which operates on the semantically structured object list. For each object, the system infers whether it causes a blind spot and, if so, describes the occlusion using a fixed output format:  
\texttt{\#ObjectID \#ObjectType has a blind spot on its \#Direction, saliency: \#High/Medium/Low, reason: \#Explanation}.

To avoid logical entanglement and unnecessary complexity in spatial reasoning, we adopt a looped querying structure in practice. That is, the system iterates over all recognized objects individually and queries the VLMs to assess whether each object may be causing a blind spot. This modular reasoning structure allows for stable and interpretable blind spot detection across varying scene layouts.

The previous blind spot analysis does not take driving intention into account. As a result, it may identify occlusions that are irrelevant to the current driving task and do not require perception completion. For instance, during highway driving in a straight lane, a vehicle in the adjacent left lane may block visibility of vehicles further left. While such occlusions would be flagged by the earlier spatial reasoning module, they do not impact the safety of a vehicle whose intention is to maintain a straight trajectory.

To address this, the \textit{Potential Risk Analysis} module integrates the inferred blind spots with the ego vehicle’s driving intention to determine whether additional perception is necessary. Specifically, if any identified blind spot—regardless of its saliency level—intersects with the intended path of travel, the agent flags a potential perception gap. In such cases, the system prepares to initiate targeted cooperative communication with relevant neighboring agents.

This layered reasoning pipeline ensures that only driving-critical blind spots trigger CP requests. It allows the system to minimize unnecessary information exchange, while maintaining safety guarantees through intent-aware filtering. The result is a more efficient and interpretable blind spot reasoning process that avoids full sensor fusion and focuses only on what matters for the current driving decision.

\subsection{Dialogue Target Matching}

Once a potential driving-critical blind spot is identified in the image space, the agent must determine which peer agent is most suitable to provide the missing information. This process consists of two tightly coupled steps: matching the occluded object in the perception image to a real-world connected agent based on spatial and visual cues, and initiating a structured dialogue to retrieve targeted information. To achieve this, we design a Dialogue Target Matching Model that leverages geometric reasoning and field-of-view (FOV) projection, followed by a dialogue direction identification step that formalizes the query content.

Given the ego agent \( \omega \in \Omega_C^t \) with state \( s_\omega^t = (x_\omega^t, y_\omega^t, \phi_\omega^t, v_\omega^t) \), driving intetion $ u_\omega^t$, static dimensions \( (l_\omega, w_\omega) \), and front-view camera parameters including horizontal FOV \( \Theta_{\mathrm{FOV}} \) and image width \( W_{\mathrm{img}} \), we consider a set of surrounding connected agents \( \mathcal{A}^t = \{\omega'_1, \omega'_2, \dots\} \subset \Omega_C^t \setminus \{\omega\} \). Each peer agent \( \omega'_j \in \Omega_C^t  \) has known center position \( (x_{\omega'_j}^t, y_{\omega'_j}^t) \), heading angle \( \phi_{\omega'_j}^t \), and dimensions \( (l_{\omega'_j}, w_{\omega'_j}) \). The driving-critical blind zone is identified in the image plane as a horizontal interval \( [u_{\min}, u_{\max}] \subset [0, W_{\mathrm{img}}] \), along with its relative occlusion direction.

To identify the most relevant peer for dialogue, we project each agent \( \omega'_j \in \Omega_C^t  \) into the ego agent’s image frame. The bearing angle from the ego agent \( \omega \) to \( \omega'_j \) is computed as
\[
\theta_j = \arctan2(y_{\omega'_j}^t - y_\omega^t, x_{\omega'_j}^t - x_\omega^t) - \phi_\omega^t.
\]
If \( \theta_j \in [-\Theta_{\mathrm{FOV}}/2, \Theta_{\mathrm{FOV}}/2] \), the peer agent is within the FOV. The corresponding horizontal pixel coordinate is
\[
u_j = \left( \frac{\theta_j}{\Theta_{\mathrm{FOV}}} + \frac{1}{2} \right) \cdot W_{\mathrm{img}}.
\]
The apparent width of the agent in pixels is approximated as
\[
w_j = \frac{w_{\omega'_j}}{d_j} \cdot f, \quad \text{where } d_j = \sqrt{(x_{\omega'_j}^t - x_\omega^t)^2 + (y_{\omega'_j}^t - y_\omega^t)^2}.
\]
Hence, the agent \( \omega'_j \) occupies an image interval \( [u_j - \tfrac{w_j}{2}, u_j + \tfrac{w_j}{2}] \). The one with the largest overlap with the blind zone \( [u_{\min}, u_{\max}] \) is selected as the matched target:
\[
\hat{j} = \arg\max_{\omega'_j \in \mathcal{A}^t} \left| [u_j - \tfrac{w_j}{2}, u_j + \tfrac{w_j}{2}] \cap [u_{\min}, u_{\max}] \right|.
\]

To determine the direction from which information is needed, we compute the relative bearing from agent \( \omega \) to the matched agent \( \omega^* = \omega'_{\hat{j}} \) as
\[
\theta_{\omega \to \omega^*} = \arctan2(y_{\omega^*}^t - y_\omega^t, x_{\omega^*}^t - x_\omega^t),
\]
and define the relative azimuth with respect to \( \omega^* \)'s heading as
\[
\Delta \theta = \theta_{\omega \to \omega^*} - \phi_{\omega^*}^t.
\]
The query direction is then discretized as:
\[
\text{QueryDirection} =
\begin{cases}
\text{"front"} & \text{if } -\tfrac{\pi}{6} < \Delta \theta < \tfrac{\pi}{6} \\
\text{"left"} & \text{if } \tfrac{\pi}{6} \leq \Delta \theta < \tfrac{5\pi}{6} \\
\text{"right"} & \text{if } -\tfrac{5\pi}{6} < \Delta \theta \leq -\tfrac{\pi}{6} \\
\text{"rear"} & \text{otherwise}
\end{cases}
\]

The final output of this module is the matched dialogue target \( \omega^* \in \Omega_C^t  \) and the corresponding directional query label \texttt{QueryDirection}. This mechanism enables the agent to initiate dialogue with a specific peer and request perception only in the occluded direction, ensuring efficient, interpretable, and low-bandwidth CP. The complete procedure is summarized in Algorithm~\ref{alg:dialogue_matching}.

\begin{algorithm}[h]
\caption{Dialogue Target Matching and Direction Identification}
\label{alg:dialogue_matching}
\begin{algorithmic}[1]
\REQUIRE Ego agent \( \omega \) with state \( (x_\omega^t, y_\omega^t, \phi_\omega^t) \), camera FOV \( \Theta_{\mathrm{FOV}} \), image width \( W_{\mathrm{img}} \), blind zone \( [u_{\min}, u_{\max}] \), peer agents \( \Omega_C^t \setminus \{\omega\} \), dimensions \( w_{\omega'} \)
\STATE Initialize \( \text{BestMatch} \leftarrow \emptyset \), \( \text{MaxOverlap} \leftarrow 0 \)
\FOR{each \( \omega' \in \Omega_C^t \setminus \{\omega\} \)}
    \STATE \( \theta_{\omega'} \leftarrow \arctan2(y_{\omega'}^t - y_\omega^t, x_{\omega'}^t - x_\omega^t) - \phi_\omega^t \)
    \IF{\( \theta_{\omega'} \in [-\tfrac{1}{2}\Theta_{\mathrm{FOV}}, \tfrac{1}{2}\Theta_{\mathrm{FOV}}] \)}
        \STATE \( u_{\omega'} \leftarrow \left( \frac{\theta_{\omega'}}{\Theta_{\mathrm{FOV}}} + \frac{1}{2} \right) \cdot W_{\mathrm{img}} \)
        \STATE \( d_{\omega'} \leftarrow \sqrt{(x_{\omega'}^t - x_\omega^t)^2 + (y_{\omega'}^t - y_\omega^t)^2} \)
        \STATE \( w' \leftarrow \frac{w_{\omega'}}{d_{\omega'}} \cdot f \)
        \STATE Projected range \( R_{\omega'} \leftarrow [u_{\omega'} - \tfrac{w'}{2}, u_{\omega'} + \tfrac{w'}{2}] \)
        \STATE \( \text{Overlap}_{\omega'} \leftarrow \left| R_{\omega'} \cap [u_{\min}, u_{\max}] \right| \)
        \IF{\( \text{Overlap}_{\omega'} > \text{MaxOverlap} \)}
            \STATE \( \text{BestMatch} \leftarrow \omega' \), \( \text{MaxOverlap} \leftarrow \text{Overlap}_{\omega'} \)
        \ENDIF
    \ENDIF
\ENDFOR
\STATE \( \omega^* \leftarrow \text{BestMatch} \)
\STATE \( \theta_{\omega \to \omega^*} \leftarrow \arctan2(y_{\omega^*}^t - y_\omega^t, x_{\omega^*}^t - x_\omega^t) \)
\STATE \( \Delta \theta \leftarrow \theta_{\omega \to \omega^*} - \phi_{\omega^*}^t \)
\IF{\( -\frac{\pi}{6} < \Delta \theta < \frac{\pi}{6} \)}
    \STATE \texttt{QueryDirection} \( \leftarrow \) "front"
\ELSIF{\( \frac{\pi}{6} \leq \Delta \theta < \frac{5\pi}{6} \)}
    \STATE \texttt{QueryDirection} \( \leftarrow \) "left"
\ELSIF{\( -\frac{5\pi}{6} < \Delta \theta \leq -\frac{\pi}{6} \)}
    \STATE \texttt{QueryDirection} \( \leftarrow \) "right"
\ELSE
    \STATE \texttt{QueryDirection} \( \leftarrow \) "rear"
\ENDIF
\RETURN Matched agent \( \omega^* \), direction \texttt{QueryDirection}
\end{algorithmic}
\end{algorithm}









\subsection{Perception Completion via Dialogue}
\label{sec:dialogue}

Once the dialogue target agent \( \omega^* \) and the query direction 
\texttt{QueryDirection} are determined, the ego agent \( \omega \) initiates a 
spatially grounded query to obtain information about the occluded region.  
The core objective of this dialogue is not to perform open-ended language 
generation, but to request only the missing perceptual content necessary 
to supplement the ego agent's current blind-spot awareness.

The dialogue mechanism follows a lightweight, interpretable structure to 
ensure consistency across agents. The ego agent issues a directional query 
anchored to the occluded zone, while the responding agent returns a structured 
description of objects, distances, and attributes relevant to the requested 
direction. After receiving the response, the ego agent fuses the reported 
object information with its local driving intention \( u_\omega^t \) to update 
risk assessment and decision-making. If the returned object intersects with or 
poses risk to the future trajectory, the ego agent adjusts its behavior (e.g., 
slowing down, yielding, or maintaining caution). This targeted interaction 
enables perception completion with minimal bandwidth and high interpretability.

\paragraph{Dialogue Prompt Format.}
The actual prompts used for querying and responding follow the patterns below.

\begin{center}
\textbf{Ego Agent Query Prompt}
\end{center}

\begin{quote}
\texttt{You are a connected vehicle agent. Based on the driving scene, you have}
\texttt{detected a potential blind spot in the [QueryDirection] direction.}
\texttt{Describe what you observe in that direction using concise object-level}
\texttt{information.}
\end{quote}

\begin{center}
\textbf{Responding Agent Output Template}
\end{center}

\begin{quote}
\texttt{There is a [ObjectType] at [RelativeDistance]}
\texttt{and its motion state is [AdditionalAttributes].}
\end{quote}

Here, \texttt{[ObjectType]} denotes object category (e.g., vehicle, pedestrian), 
\texttt{[RelativeDistance]} uses discretized spatial terms (e.g., close, medium, far), 
and \texttt{[AdditionalAttributes]} may include color, heading, or whether the object 
is stationary or accelerating.

This structured design provides a lightweight and interpretable mechanism for 
perception sharing, enabling the ego agent to supplement missing information 
in occluded regions while maintaining minimal communication overhead.

\section{Experiments}
\label{sec:experiment}

This section evaluates the proposed agentic perception framework across three core aspects: the \textbf{computational performance} of the complete perception–decision loop, the \textbf{accuracy of blind spot reasoning}, and the \textbf{communication efficiency} of dialogue-based cooperation.

\begin{figure}
     \centering
     \includegraphics[width=1\linewidth]{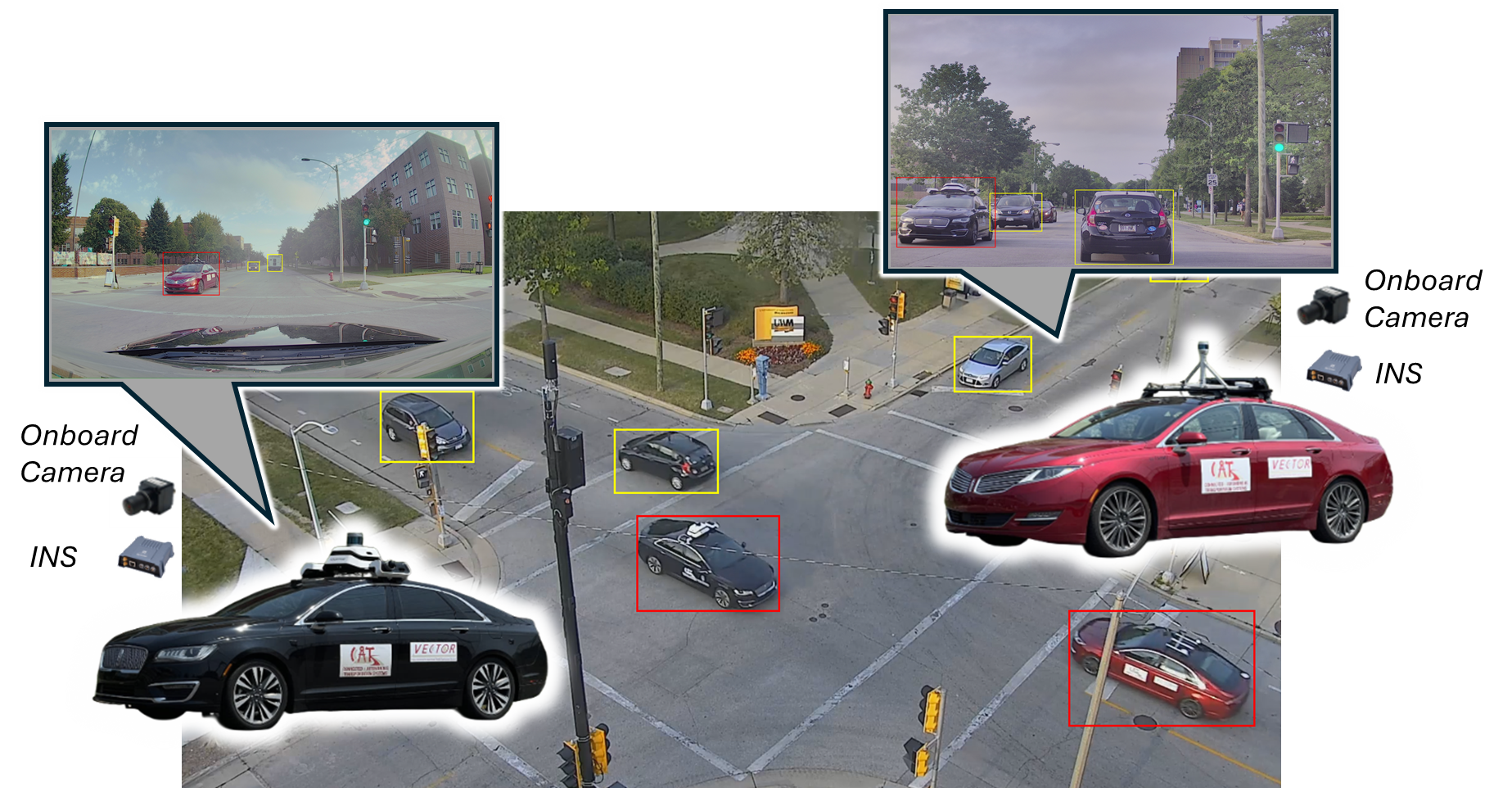}
     \caption{Data collection platform and connected vehicle experiments. }
     \label{fig:CV}
 \end{figure}

To collect real-world CP data, we utilize two CVs retrofitted by our lab, both based on identical platforms—2016 Lincoln MKZs upgraded by the University of Wisconsin-Madison. Each vehicle is equipped with a front-facing Mobileye camera with a \(120^\circ\) horizontal FOV and a Novatel OEM7 GPS navigation unit for accurate localization. All perception and communication modules are integrated on an industrial-grade onboard computer (AutonomouStuff Spectra). The vehicles operate in various urban scenarios, including signalized intersections, straight road segments, and complex work zones, as shown in Fig~\ref{fig:CV}. 

Due to the current latency limitations of LLMs, particularly during visual-language grounding and dialogue reasoning, the proposed framework is not yet deployable for real-time operations. As such, we conduct all experiments using a replay-based offline simulation pipeline. Logged sensor data and ground-truth annotations from real-world traffic scenes are replayed frame by frame, preserving vehicle states, perception inputs, and synchronized occlusion ground truth, while allowing agentic reasoning and dialogue generation to proceed under controlled conditions.

\subsection{Computation Time Analysis}

We conducted offline evaluations using a desktop equipped with an Intel i9 processor, 64\,GB RAM, and an NVIDIA RTX 4090 GPU. The analysis focuses on three key stages in our pipeline: object detection, blind spot reasoning, and dialogue-based communication between agents.

Object detection was performed using the YOLOv8 model~\citep{Jocher2023YOLO}. For blind spot reasoning, we evaluated four VLMs: DeepSeek-VL-7B, Claude-3, GPT-4o, and LLaVA-1.5-7B. Note that model initialization and loading times are excluded from our timing measurements, as these occur only once at startup and are not representative of per-frame inference latency.

\begin{figure}
    \centering
    \includegraphics[width=0.92\linewidth]{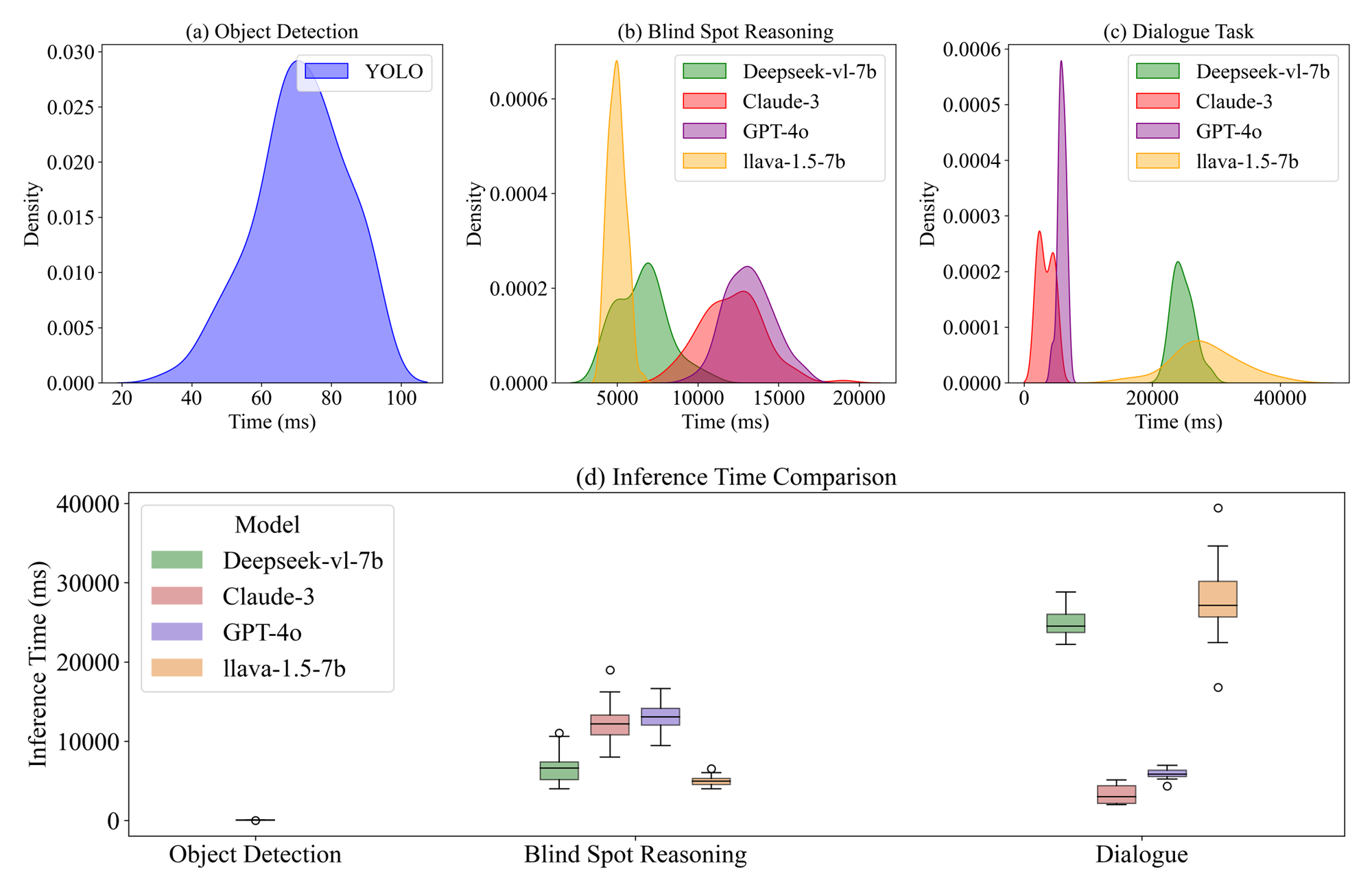}
    \caption{Computation time comparison for object detection, blind spot reasoning, and dialogue generation across different models.}
    \label{fig:cptime}
\end{figure}
Across 100 front-view frames, the computation time results are shown in Fig \ref{fig:cptime}. The YOLOv8 model completed object detection within 70\,ms per frame, providing fast and reliable detection results suitable for real-time applications. For blind spot reasoning, the average response time varied significantly across models. LLaVA-1.5-7B was the fastest, typically requiring about 5 seconds, though its reasoning quality was relatively low (as discussed in the following section). DeepSeek-VL-7B achieved a balanced performance, with a reasoning time around 7 seconds per frame. GPT-4o and Claude-3 both took longer, averaging between 12 to 15 seconds, but exhibited stronger reasoning performance.

Dialogue generation time also differed across models. LLaVA-1.5-7B and DeepSeek-VL-7B required between 20 to 35 seconds to handle multi-turn conversational inputs, largely due to their internal handling of long dialogue histories. In contrast, GPT-4o and Claude-3 responded more efficiently, demonstrating shorter response times and better suitability for agentic dialogue-driven CP tasks.

\subsection{Case Studies}

We conducted two types of case studies. The first focuses on analyzing driving-critical blind spot detection by evaluating how a single vehicle's front-view perception performs in identifying blind spots (Step 1). The second case study involves a scenario with two vehicles operating simultaneously. By comparing their individual perception results and cooperative dialogue, we analyze the full end-to-end process of driving-critical blind spot detection and correction.

For the Step 1 case study, we present two real-world scenes involving significant visual occlusions~\cite{ma2026realtime}. These cases are analyzed using the DeepSeek-VL-7B model, following object detection performed by a YOLO-based deep learning pipeline. This setup allows us to evaluate the VLM's reasoning ability in the presence of occluding objects, based on semantically enriched scene input.

\begin{figure}
    \centering
    \includegraphics[width=0.9\linewidth]{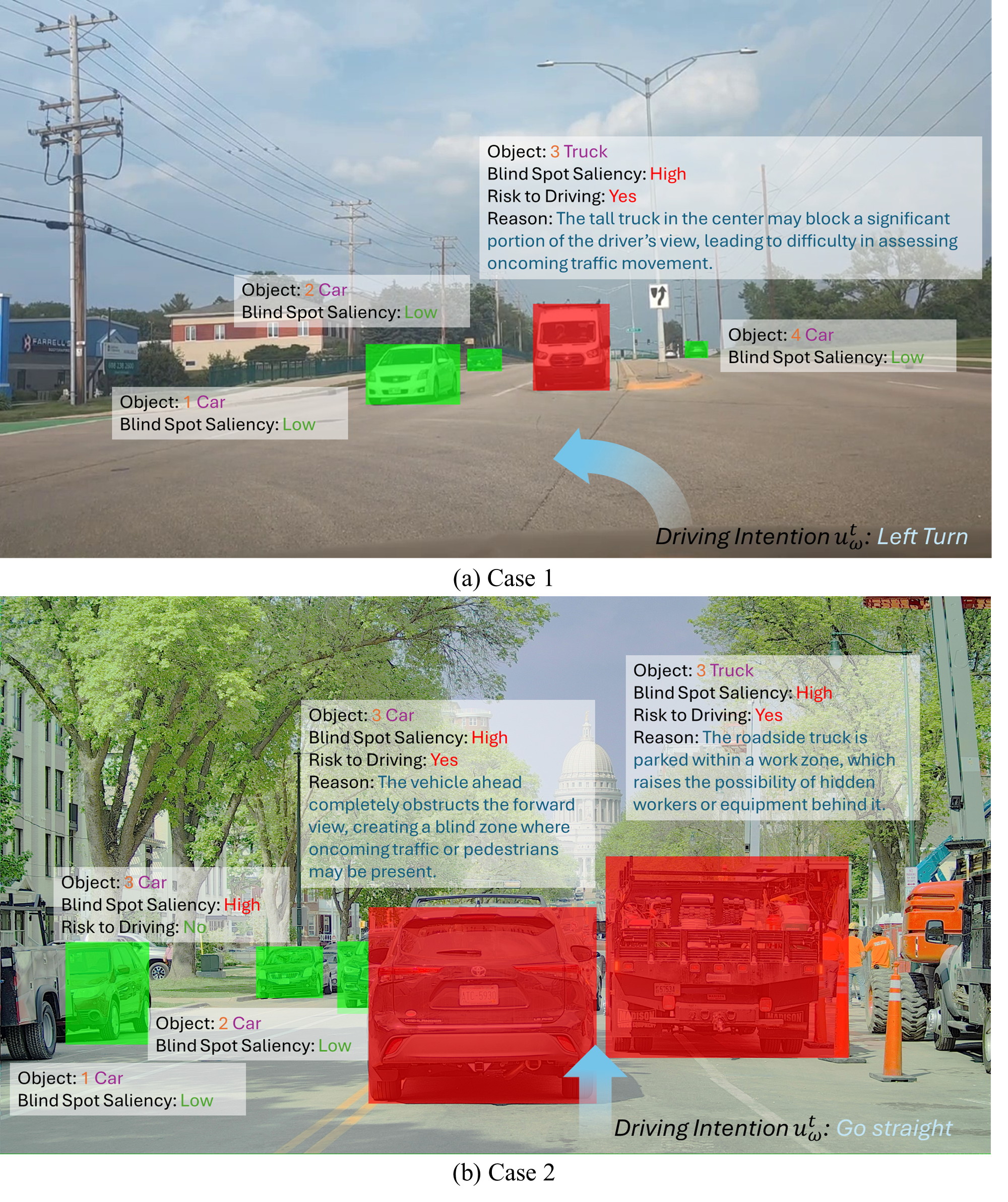}
    \caption{Case studies of blind spot reasoning}
    \label{fig:BS_reason}
\end{figure}

As shown in Fig. \ref{fig:BS_reason}, case 1 occurs at a signalized intersection where the ego vehicle intends to make a left turn. A vehicle from the opposite direction is also waiting to turn left, positioned directly in front of the ego vehicle’s line of sight. This obstructs the ego vehicle’s view of potential oncoming through traffic, making it difficult to determine a safe left-turn gap.

Without CP, such scenarios are both common and hazardous, as the ego vehicle also blocks the opposing vehicle’s view, potentially leading to a mutual deadlock. If the ego vehicle initiates a left turn without adequate visibility, it risks colliding with fast-approaching through traffic. In our framework, after detecting surrounding objects, the VLM evaluates each for its blind spot saliency—categorized as \textit{High}, \textit{Medium}, or \textit{Low}—based on their position and properties in the image. It then determines whether the resulting occlusion could impact driving decisions and provides reasoning when relevant. In this case, the left-turning vehicle from the opposing direction was correctly identified as causing a \textit{High} blind spot with potential risk. Other vehicles in the scene did not significantly obstruct the FOV and were correctly filtered out. Upon identifying the driving-critical blind spot, the ego agent initiated a dialogue with the occluding vehicle to query the visibility of the obstructed region. After receiving confirmation that no vehicles were present in the blind zone, the ego vehicle proceeded safely with the left turn. A detailed demonstration of the dynamic process is available at: \url{https://youtu.be/pgDqkrz4QT4}.

The second case in Fig. \ref{fig:BS_reason} involves a forward-driving scenario where the ego vehicle travels through a work zone area. A construction truck is parked on the roadside, partially obstructing the view of potential workers or equipment behind it. Additionally, another vehicle in front of the ego vehicle blocks a significant portion of the forward view, making it difficult to assess the full driving environment. Without CP, the ego vehicle must slow down and continuously observe the scene to ensure safety, which still leaves residual risks due to limited visibility. Using our proposed method, the system first identifies all visible objects in the scene and evaluates their potential to cause visual occlusions. Blind spot saliency is assessed, and for each significant occlusion, the model determines whether it could impact the driving task. In this case, the construction truck and the leading vehicle were successfully identified as major driving-critical blind spot sources that could interfere with safe forward movement. In contrast, other roadside vehicles and moving objects were correctly judged as not posing significant risk at the moment and were excluded from blind spot warnings. This result demonstrates the model’s capability to filter non-threatening occlusions and focus CP efforts where most needed.

\begin{figure}
  \centering
  \begin{subfigure}[t]{\linewidth}
    \centering
    \includegraphics[width=\linewidth]{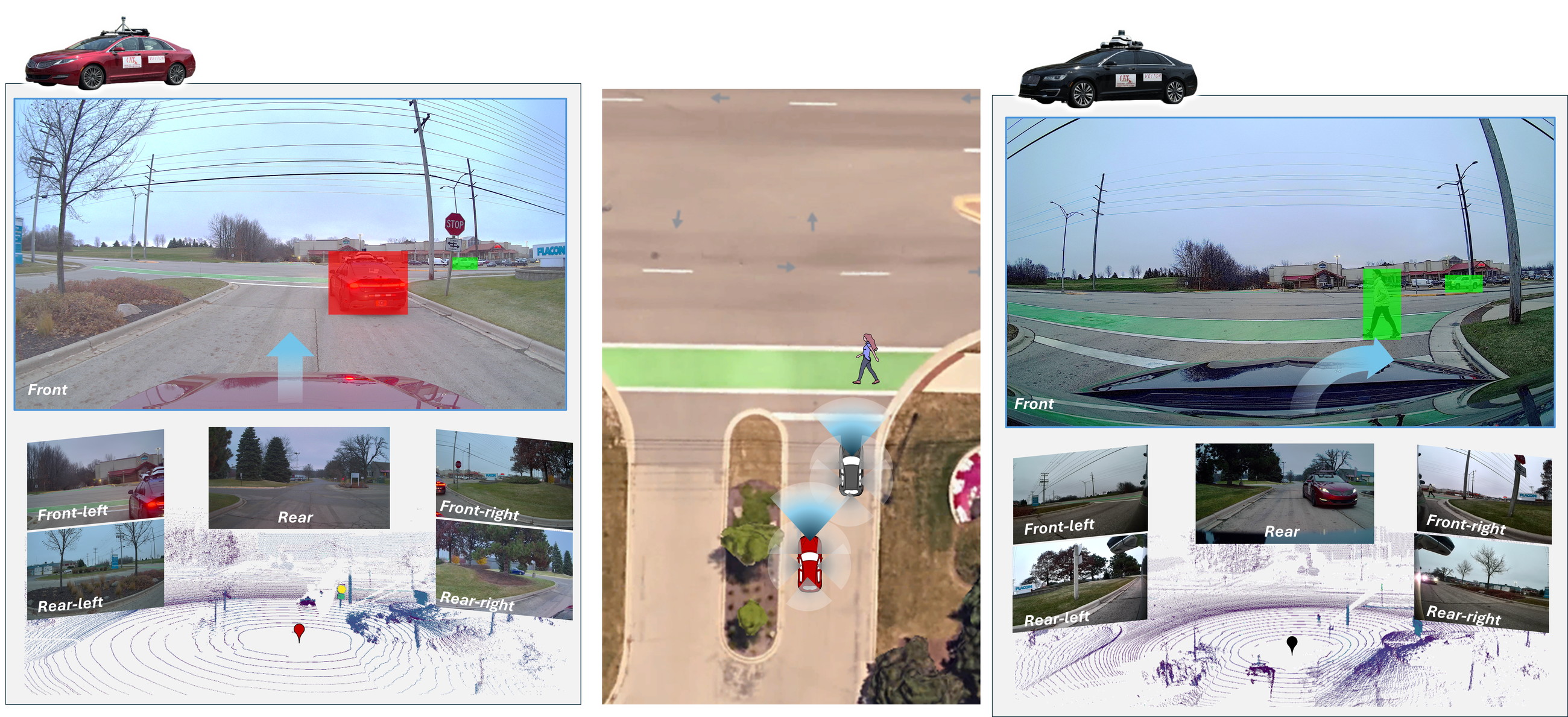}
    \caption{Case 1: a hidden pedestrian scenario where the red vehicle infers a driving-critical blind spot caused by the black vehicle and receives confirmation.}
  \end{subfigure}
  \vspace{0.3em}
  \begin{subfigure}[t]{\linewidth}
    \centering
    \includegraphics[width=\linewidth]{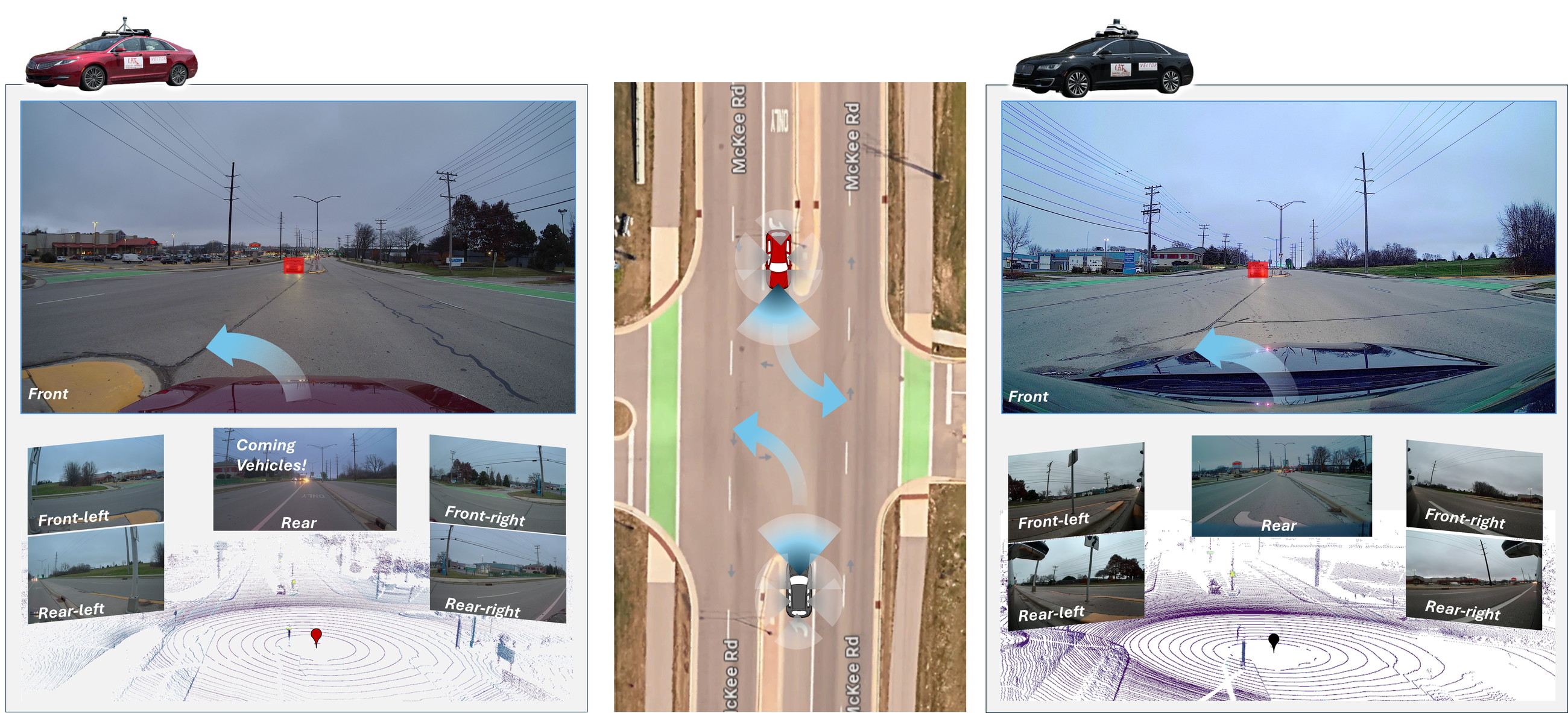}
    \caption{ Case 2: a mutual occlusion scenario at an unprotected left-turn intersection, where both vehicles identify and resolve each other’s blind spots through bidirectional communication.}
  \end{subfigure}
  \caption{Two case studies of agentic cooperative perception.}
  \label{fig:visual}
\end{figure}

The second case study focuses on the full CP process across multiple connected vehicles. As shown in Fig.~\ref{fig:visual}, we conduct end-to-end evaluation using two real-world scenes where both vehicles are simultaneously active and influence each other's observations. In Fig.~\ref{fig:visual}(a), Case 1 presents a typical \textit{“hidden pedestrian”} scenario. The center of the figure shows the bird's-eye view (BEV) of the environment, with the left and right panels displaying the perception results of the red and black vehicles, respectively. Although our framework only uses the front-facing view for blind spot detection, additional information from six-directional surround-view cameras and LiDAR is used to help interpret the scene during post-analysis (see~\cite{li2025cats} for sensor configuration details). In this scene, a pedestrian is walking in front of the black vehicle while the red vehicle attempts to pass on its side. However, the red vehicle’s view of the pedestrian is fully occluded by the black vehicle, creating a high-risk situation, especially when the pedestrian is accompanied by a child or pet. (Note: All tests were conducted in a protected environment.) 

Using the proposed Agentic CP model, the red vehicle successfully identifies the black vehicle as a \textit{driving-critical occluding object} and initiates a query: whether any vulnerable road users are present in front of it. In response, the black vehicle confirms the presence of a pedestrian currently crossing. This dialogue-based exchange allows the red vehicle to receive a proactive hazard alert and take precautionary actions, demonstrating the effectiveness of our model in blind spot correction through cooperative dialogue. The multi-source sensor data used in this scenario—including surround-view cameras and LiDAR—are processed following ~\cite{li2025cats}. The fused perception outputs are converted into structured textual descriptions and provided as input to each agent, enabling them to reason over a semantically enriched representation of the environment. This allows agents to engage in informed, multi-party dialogue to resolve blind spots and collaboratively complete the perception task.

Fig.~\ref{fig:visual}(b) illustrates Case 2, which shares the same intersection geometry as Fig.~\ref{fig:BS_reason} Case 1. In this scenario, two connected vehicles approach the same intersection from opposite directions, each attempting an unprotected left turn. From the individual perception views shown in the figure, it is evident that the two vehicles mutually occlude each other’s forward FOV. This mutual occlusion makes it difficult for either vehicle to determine whether an oncoming vehicle is proceeding straight through the intersection, often resulting in a deadlock where both vehicles hesitate to proceed.

In this case, both vehicles correctly identify that their line of sight is blocked by the opposing vehicle and determine that a driving-critical blind spot exists behind the occluding vehicle. Each initiates a query to the other, requesting information about the occluded zone.

The black vehicle replies that there is no approaching vehicle within close range behind it, indicating it is safe for the red vehicle to complete the left turn. In contrast, the red vehicle reports that a fast-approaching vehicle is coming from behind the black vehicle. It warns that initiating a left turn at that moment would pose a collision risk. This successful bidirectional exchange of critical occlusion information demonstrates the model’s capability to resolve mutual blind zones through targeted dialogue, thus breaking the deadlock and improving intersection safety.

\subsection{Acuracy of Blind Spot Reasoning}

Since no existing study has explicitly addressed blind spot reasoning via interactive dialogue in CP, we compare our method against two baselines: (1) the no-fusion strategy, where each vehicle independently perceives the environment using its own onboard sensors, mimicking human drivers; and (2) an early fusion scheme, where vehicles share raw sensor data such as images or LiDAR point clouds for centralized or decentralized joint perception. The latter represents the upper bound of bandwidth consumption but provides full visibility coverage. To evaluate the driving-critical blind spot reasoning accuracy, we compare the inferred visibility and object descriptions against ground truth, with human driver occlusion judgment used as reference. For communication efficiency, we record the average data volume exchanged per decision cycle and compare our dialogue-based selective sharing mechanism with early fusion transmission. 

In this section, we validate the effectiveness of the proposed method by evaluating the accuracy of driving-critical blind spot reasoning performed by the VLMs, as well as the success rate of the agent-to-agent dialogue process. All four VLMs introduced earlier were tested in this evaluation.

To establish ground truth, experienced human drivers reviewed the same front-view images and manually labeled key objects that caused driving-critical blind spots affecting driving decisions. These annotated objects serve as the reference for evaluating the VLMs’ blind spot identification accuracy.

The evaluation metrics for blind spot reasoning include True Positive Rate (TP Rate), False Positive Rate (FP Rate), and F1 Score. Note that we only assess the correctness of identifying which objects cause driving-critical blind spots; we do not evaluate the textual reasoning quality itself. This is because reasoning explanations, while important, are harder to quantify consistently and are not directly involved in selecting communication targets or initiating dialogue.

\begin{figure}
    \centering
    \includegraphics[width=0.8\linewidth]{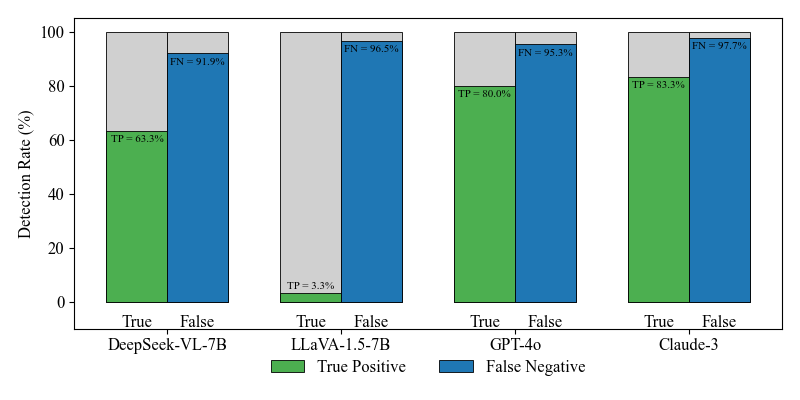}
    \caption{Driving-Critical Blind Spot Detection Accuracy}
    \label{fig:tf}
\end{figure}

The dialogue success rate is evaluated based on whether the ego vehicle successfully acquires relevant information about the blind zone—specifically, whether any critical object affecting driving was confirmed to be present or absent through the communication process. The validation results are summarized in Table~\ref{tab:validation} and Figure \ref{fig:tf} .

\begin{table}[h]
\centering
\caption{Blind Spot Detection Accuracy and Dialogue Success Rates}
\label{tab:validation}
\begin{tabular}{lcccl}
\toprule
\textbf{Model} & \textbf{TP Rate (\%)} & \textbf{FP Rate (\%)}& \textbf{F1 Score}  &\textbf{Dialogue Success Rate (\%)}\\
\midrule
DeepSeek-VL-7B & 63.3 & 8.1  & 0.727  &88.0\\
LLaVA-1.5-7B   & 3.3  & 3.5  & 0.061  &33.3\\
GPT-4o         & 80.0 & \textbf{4.7}  & 0.862  &\textbf{95.1}\\
Claude-3       & \textbf{83.3}& 2.3 & \textbf{0.893}&92.1\\
\bottomrule
\end{tabular}
\end{table}

The results show that, except for LLaVA-1.5-7B, all other models achieved reasonable performance in driving-critical blind spot identification and reasoning. Specifically, the true positive rate of blind spot recognition exceeded 60\% for DeepSeek-VL-7B, and surpassed 80\% for GPT-4o and Claude-3. The false positive rates were also relatively low, indicating a strong ability to filter out non-critical occlusions. LLaVA-1.5-7B showed weak sensitivity to this task, likely due to limitations in grounding and scene understanding. Dialogue success rates were also high across models. While DeepSeek-VL-7B exhibited slightly lower consistency in multi-turn dialogue compared to GPT-4o and Claude-3, it still performed reasonably well. Overall, the proposed approach effectively identifies driving-critical blind spot regions, significantly reducing the communication burden in CP.

\subsection{Communication Cost Analysis}

We compare the communication efficiency of the proposed agentic cooperative perception framework with conventional early fusion and late fusion approaches. Prior studies have extensively discussed the bandwidth burden imposed by early and late fusion schemes.

In early fusion methods, raw sensor data such as LiDAR point clouds or camera images are transmitted to neighboring agents or infrastructure nodes. This results in extremely high communication demands, often reaching the order of gigabits per second (Gbps)~\cite{ma2026realtime}. These methods typically require continuous high-bandwidth transmission, making them impractical under limited V2X bandwidth conditions.

Late fusion methods reduce the transmission load by only sharing high-level perception results such as detected object positions and categories. This significantly lowers the communication cost, often to the range of several kilobits per second (Kbps), depending on object count and update frequency. For example, sharing a fixed-length object list with 10–20 objects at 10 Hz typically results in data rates under 100 Kbps.

In contrast, our proposed method adopts an asynchronous, dialogue-triggered communication strategy. Vehicles broadcast only essential position and identity metadata during normal operation. Rich semantic messages are transmitted only when a driving-critical blind spot is detected and verified, which occurs infrequently—empirically less than 0.1\% of operational frames in typical urban scenes\cite{ma2026realtime}. These text messages are compact, typically consisting of 1–3 short sentences (under 500 bytes), resulting in negligible bandwidth usage even when transmitted.

Assuming a frame rate of 10 Hz and one cooperative message every 1,000 frames, the average communication load of our method is approximately:

\[
\text{Avg. bandwidth} \approx \frac{500 \text{ bytes}}{100 \text{ sec}} = 40 \text{ bps}
\]

This is three orders of magnitude lower than late fusion methods, and over six orders lower than early fusion. Therefore, our method offers a scalable and communication-efficient alternative for selective, intent-driven cooperative perception under bandwidth-constrained conditions.

\section{Conclusion}
\label{sec:conclusion}

This study proposes an efficient CP framework based on VLMs that leverages blind spot reasoning and an agentic dialogue mechanism to selectively query perceptual information from other connected agents. The framework incorporates a mixed-model architecture that combines stable object detection, vision-language reasoning, and geometric consistency, along with a CoT prompt design tailored for driving intention and occlusion assessment. We collected real-world urban traffic data using front-facing cameras and conducted comparative evaluations across four VLMs, including DeepSeek-VL-7B, GPT-4o, Claude-3, and LLaVA-1.5-7B. Results show that the proposed approach achieves over 80\% accuracy in identifying driving-critical blind spots and over 90\% success in dialogue-based CP under safety-relevant scenarios.

Future work will focus on improving computational efficiency for real-time deployment, testing in more complex and diverse traffic environments, and designing enhanced models to handle multi-agent reasoning and coordination under dynamic conditions.


\printcredits

\bibliographystyle{cas-model2-names}

\bibliography{cas-refs}

@article{li2025cats,
  title={CATS-V2V: A Real-World Vehicle-to-Vehicle Cooperative Perception Dataset with Complex Adverse Traffic Scenarios},
  author={Li, Hangyu and Cao, Bofeng and Liang, Zhaohui and Li, Wuzhen and Oh, Juyoung and Chen, Yuxuan and Liang, Shixiao and Zhou, Hang and Ma, Chengyuan and Liu, Jiaxi and others},
  journal={arXiv preprint arXiv:2511.11168},
  year={2025}
}

@misc{Jocher2023YOLO,
  author = {Jocher, Glenn and Chaurasia, Ayush and Qiu, Jing},
  title = {YOLO by Ultralytics},
  year = {2023},
  howpublished = {\url{https://github.com/ultralytics/ultralytics}}
}

@article{hou2025explainable,
  title={An explainable end-to-end autonomous driving framework based on large language model and vision modality fusion: design and application of DriveLLM-V},
  author={Hou, Junxu and Wang, Hong and Li, Tan and Dong, Eryi},
  journal={Transportation Research Part C: Emerging Technologies},
  volume={181},
  pages={105368},
  year={2025},
  publisher={Elsevier}
}

@article{ma2026realtime,
  title={Real-time identification of cooperative perception necessity in road traffic scenarios},
  author={Ma, Chengyuan and Li, Hangyu and Long, Keke and Zhou, Hang and Liang, Zhaohui and Li, Pei and Yu, Hongkai and Li, Xiaopeng},
  journal={Transportation Research Part C: Emerging Technologies},
  volume={185},
  pages={105547},
  year={2026},
  publisher={Elsevier},
  doi={10.1016/j.trc.2026.105547}
}

@misc{liu2025sracp,
  title={SRA-CP: Spontaneous Risk-Aware Selective Cooperative Perception},
  author={Liu, Jiaxi and Ma, Chengyuan and Zhou, Hang and Tang, Weizhe and Liang, Shixiao and Ding, Haoyang and Li, Xiaopeng and Ran, Bin},
  year={2025},
  eprint={2511.17461},
  archivePrefix={arXiv},
  primaryClass={cs.RO},
  doi={10.48550/arXiv.2511.17461},
  url={https://arxiv.org/abs/2511.17461}
}

@article{bai2024survey,
  title={A survey and framework of cooperative perception: From heterogeneous singleton to hierarchical cooperation},
  author={Bai, Zhengwei and Wu, Guoyuan and Barth, Matthew J and Liu, Yongkang and Sisbot, Emrah Akin and Oguchi, Kentaro and Huang, Zhitong},
  journal={IEEE Transactions on Intelligent Transportation Systems},
  year={2024},
  publisher={IEEE}
}

@article{yuan2023enhanced,
  title={Enhanced target tracking algorithm for autonomous driving based on visible and infrared image fusion},
  author={Yuan, Quan and Shi, Haixu and Xuan, Ashton Tan Yu and Gao, Ming and Xu, Qing and Wang, Jianqiang},
  journal={Journal of Intelligent and Connected Vehicles},
  volume={6},
  number={4},
  pages={237--249},
  year={2023},
  publisher={TUP}
}

@article{chiu2025v2v,
  title={V2v-llm: Vehicle-to-vehicle cooperative autonomous driving with multi-modal large language models},
  author={Chiu, Hsu-kuang and Hachiuma, Ryo and Wang, Chien-Yi and Smith, Stephen F and Wang, Yu-Chiang Frank and Chen, Min-Hung},
  journal={arXiv preprint arXiv:2502.09980},
  year={2025}
}

@article{zheng2024intelligent,
  title={Intelligent decision-making method for vehicles in emergency conditions based on artificial potential fields and finite state machines},
  author={Zheng, Xunjia and Li, Huilan and Zhang, Qiang and Liu, Yonggang and Chen, Xing and Liu, Hui and Luo, Tianhong and Gao, Jianjie and Xia, Lihong},
  journal={Journal of Intelligent and Connected Vehicles},
  volume={7},
  number={1},
  pages={19--29},
  year={2024},
  publisher={TUP}
}

@inproceedings{bai2022pillargrid,
  title={Pillargrid: Deep learning-based cooperative perception for 3d object detection from onboard-roadside lidar},
  author={Bai, Zhengwei and Wu, Guoyuan and Barth, Matthew J and Liu, Yongkang and Sisbot, Emrah Akin and Oguchi, Kentaro},
  booktitle={2022 IEEE 25th International Conference on Intelligent Transportation Systems (ITSC)},
  pages={1743--1749},
  year={2022},
  organization={IEEE}
}

@inproceedings{xu2022v2x,
  title={V2x-vit: Vehicle-to-everything cooperative perception with vision transformer},
  author={Xu, Runsheng and Xiang, Hao and Tu, Zhengzhong and Xia, Xin and Yang, Ming-Hsuan and Ma, Jiaqi},
  booktitle={European conference on computer vision},
  pages={107--124},
  year={2022},
  organization={Springer}
}

@article{bai2023cooperverse,
  title={Cooperverse: A mobile-edge-cloud framework for universal cooperative perception with mixed connectivity and automation},
  author={Bai, Zhengwei and Wu, Guoyuan and Barth, Matthew J and Liu, Yongkang and Sisbot, Emrah Akin and Oguchi, Kentaro},
  journal={arXiv preprint arXiv:2302.03128},
  year={2023}
}

@article{zhang2023building,
  title={Building cooperative embodied agents modularly with large language models},
  author={Zhang, Hongxin and Du, Weihua and Shan, Jiaming and Zhou, Qinhong and Du, Yilun and Tenenbaum, Joshua B and Shu, Tianmin and Gan, Chuang},
  journal={arXiv preprint arXiv:2307.02485},
  year={2023}
}

@inproceedings{zhang2024proagent,
  title={Proagent: building proactive cooperative agents with large language models},
  author={Zhang, Ceyao and Yang, Kaijie and Hu, Siyi and Wang, Zihao and Li, Guanghe and Sun, Yihang and Zhang, Cheng and Zhang, Zhaowei and Liu, Anji and Zhu, Song-Chun and others},
  booktitle={Proceedings of the AAAI Conference on Artificial Intelligence},
  volume={38},
  pages={17591--17599},
  year={2024}
}

@article{du2024constrained,
  title={Constrained Human-AI Cooperation: An Inclusive Embodied Social Intelligence Challenge},
  author={Du, Weihua and Lyu, Qiushi and Shan, Jiaming and Qi, Zhenting and Zhang, Hongxin and Chen, Sunli and Peng, Andi and Shu, Tianmin and Lee, Kwonjoon and Dariush, Behzad and others},
  journal={arXiv preprint arXiv:2411.01796},
  year={2024}
}

@article{xu2024vlm,
  title={Vlm-ad: End-to-end autonomous driving through vision-language model supervision},
  author={Xu, Yi and Hu, Yuxin and Zhang, Zaiwei and Meyer, Gregory P and Mustikovela, Siva Karthik and Srinivasa, Siddhartha and Wolff, Eric M and Huang, Xin},
  journal={arXiv preprint arXiv:2412.14446},
  year={2024}
}

@article{jiang2025survey,
  title={A Survey on Vision-Language-Action Models for Autonomous Driving},
  author={Jiang, Sicong and Huang, Zilin and Qian, Kangan and Luo, Ziang and Zhu, Tianze and Zhong, Yang and Tang, Yihong and Kong, Menglin and Wang, Yunlong and Jiao, Siwen and others},
  journal={arXiv preprint arXiv:2506.24044},
  year={2025}
}

@article{huang2025survey,
  title={A survey on hallucination in large language models: Principles, taxonomy, challenges, and open questions},
  author={Huang, Lei and Yu, Weijiang and Ma, Weitao and Zhong, Weihong and Feng, Zhangyin and Wang, Haotian and Chen, Qianglong and Peng, Weihua and Feng, Xiaocheng and Qin, Bing and others},
  journal={ACM Transactions on Information Systems},
  volume={43},
  number={2},
  pages={1--55},
  year={2025},
  publisher={ACM New York, NY}
}

@article{tian2025nuscenes,
  title={Nuscenes-spatialqa: A spatial understanding and reasoning benchmark for vision-language models in autonomous driving},
  author={Tian, Kexin and Mao, Jingrui and Zhang, Yunlong and Jiang, Jiwan and Zhou, Yang and Tu, Zhengzhong},
  journal={arXiv preprint arXiv:2504.03164},
  year={2025}
}

@article{tang2024sparkle,
  title={Sparkle: Mastering basic spatial capabilities in vision language models elicits generalization to composite spatial reasoning},
  author={Tang, Yihong and Qu, Ao and Wang, Zhaokai and Zhuang, Dingyi and Wu, Zhaofeng and Ma, Wei and Wang, Shenhao and Zheng, Yunhan and Zhao, Zhan and Zhao, Jinhua},
  year={2024}
}

\end{document}